# Learning Representative Temporal Features for Action Recognition

**Ali Javidani[*,**], Ahmad Mahmoudi-Aznaveh[**]**

[*] Department of Electrical and Computer Engineering, Queen's University, Kingston, Canada

[**] Cyberspace Research Center, Shahid Beheshti University, Tehran, Iran

***Abstract-*** In this paper, a novel video classification method is presented that aims to recognize different categories of third-person videos efficiently. Our motivation is to achieve a light model that could be trained with insufficient training data. With this intuition, the processing of the 3-dimensional video input is broken to 1D in temporal dimension on top of the 2D in spatial. The processes related to 2D spatial frames are being done by utilizing pre-trained networks with no training phase. The only step which involves training is to classify the 1D time series resulted from the description of the 2D signals. As a matter of fact, optical flow images are first calculated from consecutive frames and described by pre-trained CNN networks. Their dimension is then reduced using PCA. By stacking the description vectors beside each other, a multi-channel time series is created for each video. Each channel of the time series represents a specific feature and follows it over time. The main focus of the proposed method is to classify the obtained time series effectively. Towards this, the idea is to let the machine learn temporal features. This is done by training a multi-channel one dimensional Convolutional Neural Network (1D-CNN). The 1D-CNN learns the features along the only temporal dimension. Hence, the number of training parameters decreases significantly which would result in the trainability of the method on even smaller datasets. It is illustrated that the proposed method could reach the state-of-the-art results on two public datasets UCF11, jHMDB and competitive results on HMDB51.

## 1. Introduction

Human Activity Recognition (HAR) is one of the popular research areas in artificial intelligence. The industrial applications of this field are surveillance systems, video game consoles, driver-less cars and so forth. Because of its diverse applications, many researches have been conducted in recent years. It is also still an open area for research due to the difficult and challenging datasets. Especially those containing several moving objects with cluttered backgrounds.

In general, the input of an action recognition system is a 3D video signal consisting two dimensions in the spatial and one dimension in the temporal domain. The output of such system is the category that the video belongs to [1]. As a matter of fact, this problem is the extension of the image classification in which frames of the videos are placed over the temporal dimension continuously. However, the challenges of the video classification are not comparable with the image recognition [2, 3]. In videos, not only the spatial algorithms have to be performed to analyze each frame of the video separately, but also, they should be followed up over time to find the relations between consecutive frames. The processes of the later one lead to serious overweight in time and space complexities[2, 4, 5]. As a result, by looking at the video classification problem generally, it is expected that complex models with huge number of training parameters are required to perform the task with sufficient accuracy [1, 2, 4-11].

In videos, unlike images, the main point is to take the advantage of both the spatial and the temporal streams to extract spatio-temporal features for representing the video data efficiently [1, 4, 12, 13]. Before the genesis of deep learning, many image domain algorithms were either directly applied or extended to three dimensions to be used for the video classification. The procedure was as follows. First, the key points of each frame were extracted, either densely or sparsely, as interest points[14, 15] . These key points were described by common descriptors such as HOG [16] and Sift [15]. In this case, HOF and MBH were mainly used since they could simulate the existing motion in videos [17-19]. Also, many of these descriptors were extended to three dimensions to incorporate the temporal dimension in order to capture dynamic changes more effectively [20, 21]. Then, by performing encoding representation methods, the video-level features were extracted from the frame-level information. Bag of Visual Words (BoVW) [22], Fisher Vector (FV) [23, 24], and VLAD [25] were amongst the most prevalent algorithms. As the last step, the video-level extracted features were fed to a classifier to categorize them [26, 27].

The aforementioned pipeline was highly explored for many years. One of its breakthroughs was the improved Dense Trajectory (iDT) method which could exhibit outstanding results [28, 29]. However, these traditional methods were being suffered by some inherent problems. As an example, each hand-crafted descriptor notices only one aspect of the video frame to extract advantageous features from it. While, there may exist many aspects where considering all of them would be impossible. Descriptors, even their 3D extensions, are not capable of capturing long-term temporal relations. They can only capture short-term ones which are limited between adjacent frames. Furthermore, most of these methods are highly time and space consuming. For instance, the clustering step, which is the indispensable phase of Bag of Visual Words (BoVW), requires relatively high amount of time and also space to be performed completely.

With the invention of Convolutional Neural Networks (CNNs), the research direction has been entirely changed . It is clearly illustrated that the CNNs are able to outperform the existing feature extraction methods in many applications such as image classification, digit recognition, object detection and so on [3]. Hence, it is expected that the impressive results should happen for videos as well. While, due to the fact that the CNN's inputs have been designed mostly for 2D inputs, there are some major challenges in utilizing them for videos. Firstly, how to feed the 3D video signal to 2D CNN model? There are many ways to do this, but each of them has its own problems. For instance, video frames can be processed either individually or unitedly. The former scenario can be done by feeding each frame to the CNN and achieving its feature map. In this way, combining these feature vectors together to construct video-level features is non-trivial [30-32]. On the other hand, processing some frames in a group requires to change the convolution operation to either 3D or multi-channel 2D which would result in the explosion of number of training parameters [2, 5, 7, 9, 33-35]. The second issue is that CNNs are not able to handle variable-length inputs [1, 4]. This assumption is also another issue which requires proper solutions. These problems were two of the most important challenges in utilizing CNNs for videos.

In the above, an overview of traditional Activity Recognition methods along with some of their challenges were introduced. The rest of the paper is organized as follows. In section 2, the related works with the focus of methods using deep learning and their strategies are reviewed. Section 3, explains the proposed method and the idea behind that thoroughly. The implementation details, experimental results and a comparison with the state-of-the-art results are presented in section 4.

## 2. Related Works

A lot of studies applied 2D-CNNs for the case of videos. In [2], four models with different fusion methods are introduced. To capture temporal relations, a limited number of frames are given to different channels of CNN. However, single frame and slow fusion models have approximately similar performances. Hence, it can be concluded that the slow fusion model, in spite of its huge training time, could not detect the existing motion in videos. Zisserman *et al.* devised the two-stream network to take two different aspects of appearance and motion in videos [1]. Meanwhile, the number of input optical flow images in the flow-net had to be constrained to a limited number due to the increasing number of training parameters. The selection of these optical flow images among all frames is also another challenge. These early two research works construct the backbone of other researches in this domain.

Two different but similar studies extended CNN to 3D-CNN [5, 35]. In [35], the convolution operation was extended to 3-dimensions. Similarly, Tran *et al.* designed C3D network, in which both convolution and pooling operations have been extended to 3-dimensions [5, 11]. The main drawback of extending convolutions to 3D filters is the enlargement of training parameters which highlights the necessity to more training data so that the deep network trains properly. A combination of hand-crafted and deep-learned features (trajectory deep pooled features) demonstrated a strong representation to categorize third-person videos [36]. In their implementation of the two-stream ConvNet, they randomly select only 25 frames among the whole video which does not seem efficient. Another work considered several fusion methods of spatial and temporal streams in different places of the two-stream model and obtained better performances in compare with its baseline model [4]. However, the critics mentioned about the baseline model are still remained for this model as well.

As Recurrent Neural Networks (RNNs) are appropriate to find long-term patterns from sequential data, several researches investigated RNNs, especially Long-Short Term Memory (LSTM) [37-41]. LSTMs, due to their simple structure and lesser number of training parameters in compare with CNNs, were not able to efficiently recognize the ongoing action from videos. This leads researchers to investigate more complex RNN models such as Attention LSTMs [42-44]. In fact, these approaches were more successful. Apart from using LSTMs for keeping track of features, some methods applied them to train the parameters inside their model. For instance, Piergiovanni *et al*. and Liu *et al.* trained their temporal attention filters with the aid of LSTMs for classifying first and third-person videos [44, 45]. Deep residual networks, either recurrent or convolutional networks, were also examined by researchers [13, 40, 46]. Gammulle *et al.* proposed a multi-stream deep fusion framework in which by taking advantage of CNNs and LSTMs, spatial and temporal features were extracted respectively. It surpasses the state-of-the-art results on relatively small datasets [47].

End-to-End learning is another direction where many researches have been successful in this area [8, 38, 48, 49]. In [43, 50], an architecture is proposed in which with the aid of RNNs as encoders, they were able to train their model in an end-to-end fashion. However, despite their claims, using optical flow images means they have previously computed them which conflicts with the end-to-end definition. [30] presented a novel idea for generating optical flow images by looking it as an image reconstruction problem. The authors could succeed in improving the results of TSN based fusion [31]. There are also some works which either tried to enhance the quality of motion or learn to represent it in videos. Wang *et al.* utilized autoencoders[12] and Sun *et al.* introduced Optical Flow guided Feature (OFF) which is very fast and robust [51]. With the idea of omitting human interference, [52, 53] introduced several network architectures to learn the existing motion instead of computing flow images. Learning the motion, instead of computing it using existing methods, leads to better performances. This is due to the fact that the neural network sees the effect of different motion parts based on the given labels. Hence, it will finally understands where to put stress and increase its weights. The mentioned framework will achieve higher performances since it can detect motions specifically for each application.

Another recent work, namely I3D, combined 3D based models into the two-stream architecture and by pre-training their models with ImageNet and Kinetics datasets, they could surge the results dramatically [54]. The authors in [7], extended the I3D method with the difference that they suggested to use a single stream 3D DenseNet architecture. In fact, temporal 3D ConvNets were devised to combine temporal information across variable depths [7]. The other remarkable contribution of this work is the supervised transfer learning technique. It is also indicated that pose estimation can significantly help the video classification problem. [55] performs a human pose estimator to extract heatmaps for human joints in each frame. Then, temporally aggregating the maps and combining this method with I3D leads to high performances [55]. As a complement to the recent 3D CNNs e.g. I3D, [6] introduces a temporal pose convolution to aggregate the spatial poses throughout the frames. The multi-stream fusion in this approach achieves the state-of-the-art results. As can be seen, the very recent research direction in this domain is to exploit and utilize secondary information content e.g. human poses during performing different actions, or to recognize classes from skeleton data. However, the use of subsidiary information for recognizing actions makes the comparisons unfair, since other methods could surge their results by adding such content.

In the above, a review on the important deep-learning based methodologies on Human Action Recognition was provided. Considering the fact that these methods were slowly converged to each other through recent years, a major drawback for most of them is that they require a huge amount of data to be trained properly [1, 2, 4-11]. However, the assumption of existing vast amount of labeled data in many applications can be impractical. This paper tries to take a step toward the mentioned problem. The main motivation of our work is to exploit efficient and representative temporal features in the presence of *insufficient training data*. The proposed framework is based on converting inputs to a multi-channel time series representing the motion in video. To extract discriminant features from the time series, our strategy is to train a one dimensional Convolutional Neural Network (1D-CNN). Thanks to the reason that the 1D-CNN learns features through the only temporal dimension, the number of training parameters for our model decreases considerably. This causes the fact that our method would be able to be trained on relatively small datasets. By evaluating the proposed method on two public datasets UCF11 and jHMDB, it is demonstrated that our method outperforms others with the aid of efficient temporal representation.

## 3. Proposed Method

In this section, the theory behind our proposed scheme will be discussed completely. Fig. 1 demonstrates the proposed method pipeline in detail. First, the optical flow images between consecutive frames are calculated. These flow images are fed to pre-trained CNN networks and the resulted feature maps (before the fully-connected layers) are extracted. Then, their dimensions are reduced by performing the PCA algorithm. The descriptions for successive flow images are aligned beside each other which will result in a multi-channel time series. As the last step, the obtained time series are being classified by performing the effective one dimensional Convolutional Neural Networks (1D-CNN).

### 3.1 Optical Flow and its Description

Optical flow is a traditional and strong method for representing the existing motion in video. An optical flow image between two frames in video represents how much each pixel has been moved from one frame to the other. The most important advantage of utilizing optical flow is that it detects motion in short distances throughout the nearby frames. Therefore, fixed objects with no substantial impact in determining the motion will be eliminated. This is really crucial in such applications like recognizing activities in which motion plays the main role. Accordingly, in order to estimate short-term motion, at the first step, optical flow images between consecutive frames are computed for all training videos. From now on, the general idea is to pursue the changes of short-term motions throughout the video, so that we can detect the long-term motions. Thus, describing the resulted optical flow images with a specific set of features and following each element over time can be a major step toward this goal. In our framework, we extract off-the-shelf features from the last fully-connected layer (before soft-max) from pre-trained CNN networks. These CNN networks were previously trained on large-scale image datasets. As a result, a sequence of $n$ dimensional vector for each optical flow image is obtained; where $n$ is the number of neurons in the related feature map layer.

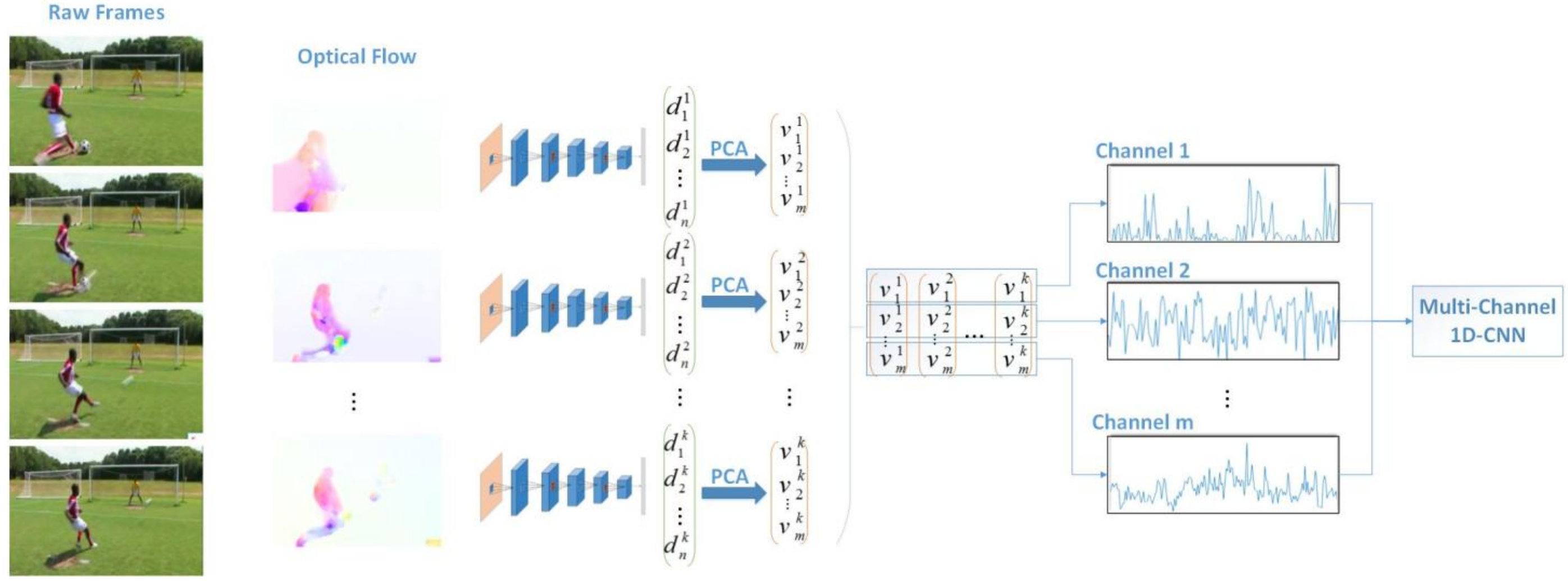

*Figure 1 The detailed structure of the proposed framework. The optical flow images between consecutive frames are calculated and described by a well pre-trained CNN network. In order to reduce the dimensionality, PCA dimension reduction algorithm is performed. Then the resulted time series are represented and classified by training a multi-channel one dimensional Convolutional Neural Network (1D-CNN). The 1D-CNN, with the aid of hierarchical representation of data, is able to find long-term temporal patterns which are useful in recognizing the ongoing action.*

### 3.2 Dimension Reduction

Optical flow images are much similar to each other. Because of that, there exist strong correlations between descriptions of them in the $n$ dimensional space. Besides, the number of dimensionalities of the last fully-connected layer in pre-trained CNN networks is usually high (e.g. 4096 in AlexNet and VGG-Net or 2048 in ResNet-152). Hence, redundancy is highly probable to occur. To omit the correlation between the optical flow images and reducing the number of dimensions, we decided to perform the simple but effective dimension reduction algorithm of Principal Component Analysis (PCA).

Principal Component Analysis (PCA) aims to find the axes exhibiting the greatest variation for projecting all data points. This is possible by finding eigenvectors of covariance matrix in the original space. By performing it, the description vectors of all optical flow images are transferred to the new space, based on the founded transformation matrices. However, the value of preserving energy should be chosen carefully. As a matter of fact, the value of Proportion of Variance (PoV) has to be selected based on considering various parameters such as the number of training data, the amount of existing correlation between the optical flow frames in the database and so on.

Looking at the problem generally, performing the dimension reduction algorithm provides two important advantages. Firstly, the volume of data that we are confronted with will be extremely lightened. This causes the fact that the data are transformed to the new space with much lesser number of dimensions, while keeping as much useful information as it can. Hopefully, in the new space the class label of an unlabeled data can be detected more deterministically. Moreover, it makes the situation more comfortable to pursue our following idea (Multi-Channel 1D-CNN) for classifying the created data. As a result, by performing the dimension reduction, all $n$ dimensional description vectors are transformed to the new space with $m$ dimensions; where $m$ is the number of dimensions in the new space and $m < n$.

Aligning the mapped descriptions of consecutive optical flow images beside each other, results in a matrix. It is clear that each row in this matrix comes from a specific feature and has been followed up over time through the columns. In other words, a set of $m$ time series are obtained which are representing motion for each video. All of these $m$ time series have their own role in classifying the whole set and they could be seen as different channels of a time series. Therefore, the problem is summarized to *multi-channel time series classification*.

### 3.3 Multi-Channel One Dimensional Convolutional Neural Network (1D-CNN)

Here, the idea is to keep track of short-term temporal features to obtain long-term patterns from the time series to classify them efficiently. In time series, short-term features can be obtained from handcrafted predefined properties like *max*, *sum* and other pooling operators [32, 56]. However, a better alternative is to let the machine learn and recognize these features. In this way, the machine learns to find those features that can optimize the classification results based on them.

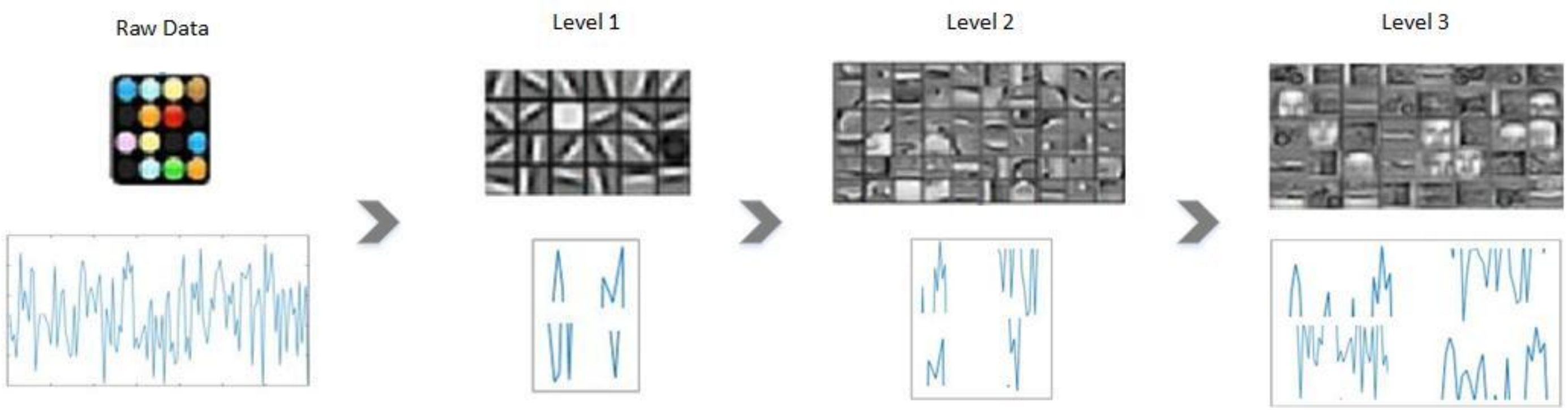

*Figure 2 Image Vs. Time Series CNNs' learnt Filter visualization. The first row is related to CNN's learnt features for images and the second row is for time series. The first column is the raw data which are fed to CNNs as inputs. The second column illustrates the first level CNNs learnt features which are like Gabor filters in images and peak and dips in time series. It is obvious that the more we go through CNN's last layers (Level 2,3), the more complex set of features are learnt e.g. human face like features in images and long-term temporal features with multiple peaks and dips in time series.*

Two properties make CNNs suitable for our purpose. Firstly, with the aid of simple back-propagation algorithm, CNNs accomplish the feature learning task automatically. Secondly, they represent the input data hierarchically. In other words, in the CNN's initial layers, some basic features are extracted. These basic features in images are equivalent to simple patterns such as Gabor like filters (first row of level 1 in Fig. 2). In time series, they are local maximums (peaks) and minimums (dips) (second row of level 1 in Fig. 2). These features can be interpreted as *sub-events of the ongoing action*. Then, the features learned in the consecutive layers are obtained by further processing of the basic features learnt in the primary layers. These complex features in images can be schematic faces of humans (first row of level 3 in Fig. 2); while in time series they consist of more complicated patterns including various peaks and valleys (second row of level 3 in Fig. 2). Accordingly, the features which are learned in the final layers represent the *super-events of actions* in the video. Hopefully, by training the Multi-Channel One Dimensional Convolutional Neural Network (1D-CNN), we are able to detect the *sub-events* and *super-events* of the progressing action in videos elegantly.

Fig. 3 demonstrates the generic form of a multi-channel 1D-CNN which consists of several 1D convolution and pooling layers. Since the input signals are one dimensional time series, all convolution and pooling operations in our Multi-Channel One Dimensional Convolutional Neural Network (1D-CNN) are defined 1D. Time series, resulted from the previous steps, are being fed to the Multi-Channel 1D-CNN. In the convolution layer, multiple 1D convolution filters are defined. These filters, by multiplying to the time series, construct feature maps which try to capture the important properties of their input. Then, max pooling layer computes the local maximum along the defined window. The convolution and pooling operations are repeated until a discriminative feature vector is extracted from the time series so that at the end, the fully-connected layers can classify them with high accuracy. It is worth to mention that everything in this network is trained with the back-propagation algorithm. In fact, based on the time series' labels, the neural network will learn the optimized convolutional kernels which results in a discriminative feature vector.

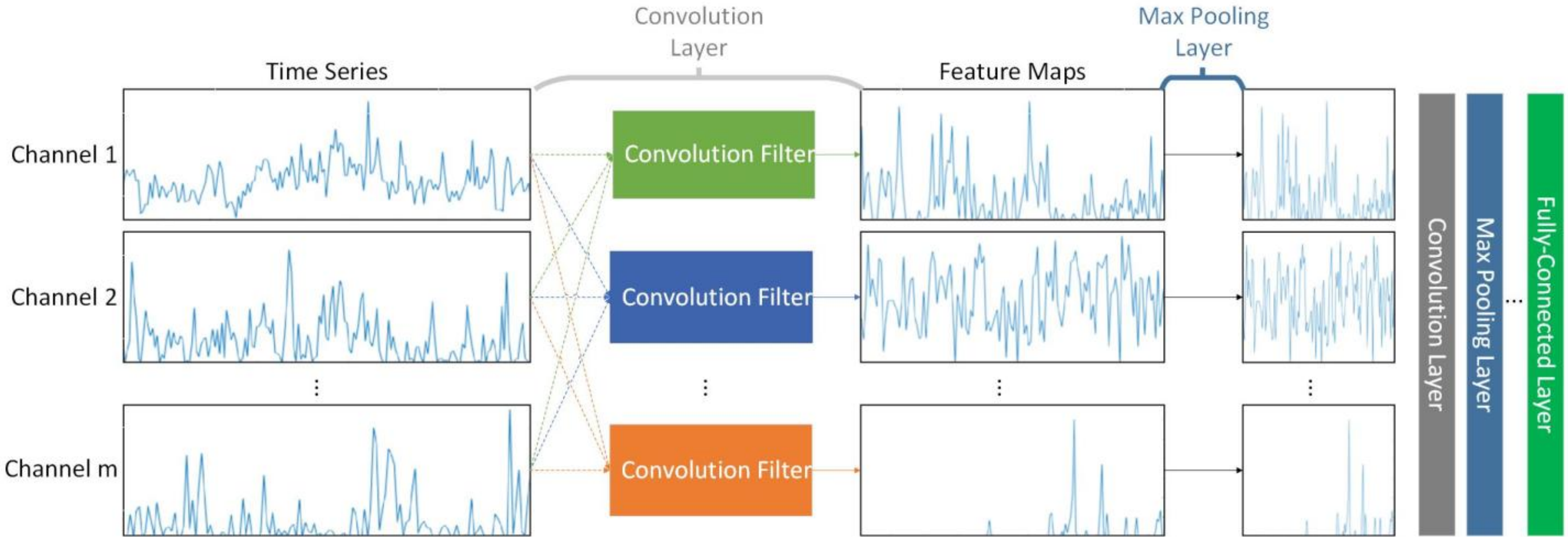

*Figure 3 The structure of Multichannel One Dimensional Convolutional Neural Network (1D-CNN). The input is a time series with m channels. Several convolutional filters are placed to learn the structure of time series to extract basic temporal features. Then, max pooling layer reduces the size of the resulted feature maps. The structure of convolution and pooling layers are repeated so that effective temporal features are extracted. Lastly, the fully connected layers are placed to classify the resulted feature vector.*

Defining the convolutional filters as one dimensional operations, will ease the training phase of the network due to the fact that the number of learning parameters will be decreased significantly. To illustrate it more, let us compare the number of training parameters between our model and the famous C3D network. The number of training parameters for a typical convolutional filter in the C3D network is $w \times h \times c$; where $w$ and $h$ are the width and the height of the filter respectively and $c$ is the same as the depth of the convolutional filter incorporating the time dimension. Whereas, in our 1D-CNN network the number of training parameters is the same as the size of the convolution filter ($c$ in the C3D). Actually, the number of training parameters has been decreased at the order of $w \times h$ for each convolutional filter. Multiplying this value by the total number of filters in the network leads to a huge number. This indicates that the number of training parameters in our model has been decreased dramatically in comparison with the C3D. Therefore, our proposed method has much more flexibility to be trained with smaller datasets. While, C3D needs massive amount of training data to train the millions of parameters inside it. This is one of the most prominent advantages of our approach in compare with other deep-learning based approaches. In the meanwhile, the proposed method is scalable too. In other words, the model's complexity can be tuned by adding more convolution and pooling layers. As a result, if there exist sufficient amount of training data, the 1D-CNN will be able to make the most out of the available data to be trained perfectly.

One of the challenges that we were confronted during the design of the 1D-CNN was that the CNNs' inputs should have a specific constant size. This is in contradiction with the problem definition of Human Activity Recognition, since the videos should have different lengths. A simple way to address this problem is to resample all videos in order to have a fixed length. However, this may result in losing useful information (specifically, when the length of the longest and the shortest videos of a database differs a lot). In the following, two premier alternatives are presented and in the experimental results section their efficacy are investigated.

### 3.4 Zero Padding Time Series

In order to train our one-dimensional Convolutional Neural Network (1D-CNN), one alternative is to set the length of each time series equal to the length of the video with maximum number of frames. In this situation, all videos having less frame numbers are expanded by zero padding. It is also worth to consider that, as our experiment results show, it does not make any difference between adding zero or another numeral to the end of the time series. This is because our Multi-channel 1D-CNN is intelligent enough to understand that repeating a constant number (either zero or any other numeral) at the end of the time series, cannot help it to distinguish between either of the two classes. Although this heuristic works well, the idea of zero padding the existing time series may not seem completely logical since adding zeros to the time series is equal to extending the videos of database with some ambiguous frames at the end of each video clip.

### 3.5 1D-CNN with Temporal Pyramid Pooling (TPP)

To address the problem of variable video lengths in a superior way, the Temporal Pyramid Pooling idea is incorporated, which is inspired by Spatial Pyramid Pooling Networks [57]. The Spatial Pyramid Pooling networks are created to solve the problem of fixed-length CNN's inputs. As it is discussed in [57], the convolution and pooling operators, which are placed at the initial layers of each CNN network, do not have any problem with variable-length inputs. The only layer that should be fed with a fixed-length input is the first fully-connected layer which is placed after the primary convolution and pooling layers. To overcome this problem, the last pooling, which is connected to the first fully connected layer in the network, is converted to a pyramid pooling.

Pyramid pooling is performed at multiple levels. The first level resembles the ordinary pooling which is done on the resulted feature map as a whole. The second level divides the feature map to $s$ equal parts and does the pooling operation on each of the sections separately which would result in $s$ dimensions. The third level is the same as the previous level with the difference that instead of $s$ sections, the feature map is divided to $s^2$ equal sections. Finally, the outputs of each level are concatenated and fed to the fully-connected neurons. In this way, the input of the fully-connected layer becomes fixed-length which is equal to $(s^0 + s^1 + s^2 + \cdots + s^L)$ dimensions; where $L$ is the number of pyramid levels.

Fig. 4 illustrates the structure of the Temporal Pyramid Pooling. Our temporal pyramid pooling structure is done with $s = 2$, $L = 2$. With these settings, in the first level of the temporal pyramid (level 1 in Fig. 4), the pooling operation is done on the time series resulted from being passed by primary convolution and pooling layers. Therefore, the result is one dimension for each of the time series. In the second level, the resulted time series is divided to two equal parts and the pooling operation is done on each of them separately which would result in two dimensions (level 2 in Fig. 4). The third level, divides the time series into four equal sections and by performing pooling operations, four dimensions would be created (level 3 in Fig. 4). Concatenating the resulted dimensions beside each other, makes the input of the fully-connected layer a fixed-length vector (7-dimensional vector for each time series).

*Figure 4 Temporal Pyramid Pooling Structure. The input time series is passed by the primary convolution and pooling operations and K feature maps are resulted. Then each of these K feature maps are being processed by the Temporal Pyramid Pooling layer. The first level takes the maximum of the whole time series. The second level halves the time series and the third level breaks the sequence into four sections, and takes the maximum across each section.*

## 4. Experimental Results

### 4.1 Datasets

To clarify the superiority of the proposed method over other deep-based approaches, which demand large-scale training data, we conducted our experiments on two rather small datasets UCF11, jHMDB. Moreover, to compare the proposed method with other state-of-the-art approaches HMDB51 was also added to our experimental reports. UCF11 contains 1600 videos distributed in 11 different sport categories of humans such as diving, volleyball and biking. The videos are collected from YouTube and the frame rate for all of them is 29.97 fps. Due to the huge inconsistency in camera motion, different backgrounds and illumination conditions, UCF11 is a challenging dataset. HMDB51 dataset contains 6766 video clips distributed in 51 classes. The classes are the daily human activities such as climbing stairs, clapping, running and so forth. Moreover, the speed of body parts such as head, legs and arms vary in different clips in this database. This dataset has 3 splits and the evaluation is based on the mean classification accuracies for those splits. JHMDB is the smaller scale of HMDB51 with only 21 out of 51 classes. The length of frames in this dataset ranges from 15 to 40. JHMDB consists of 923 videos from which around 70% is designed for training and the rest is for testing. Inheriting the above-mentioned properties from HMDB51 with considerably lesser number of video clips makes this dataset strictly challenging for categorizing. Fig. 5 illustrates sample video frames from different classes of UCF11 and HMDB51.

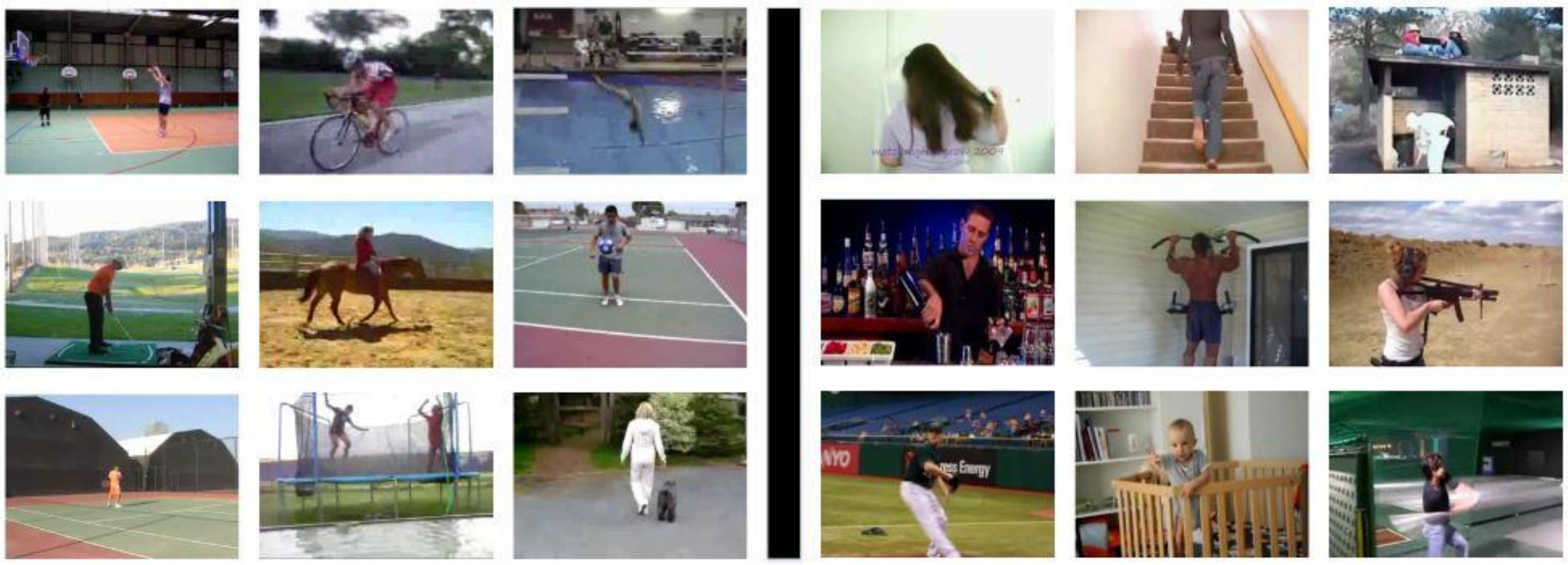

*Figure 5 Some sample frames of different classes of datasets UCF11 (left) and HMDB51 (right). As can be seen, the classes in UCF11 (left) consist of various sports such as volleyball, biking, baseball and in HMDB51 are human daily activities such as brushing hair, climbing stairs and drinking.*

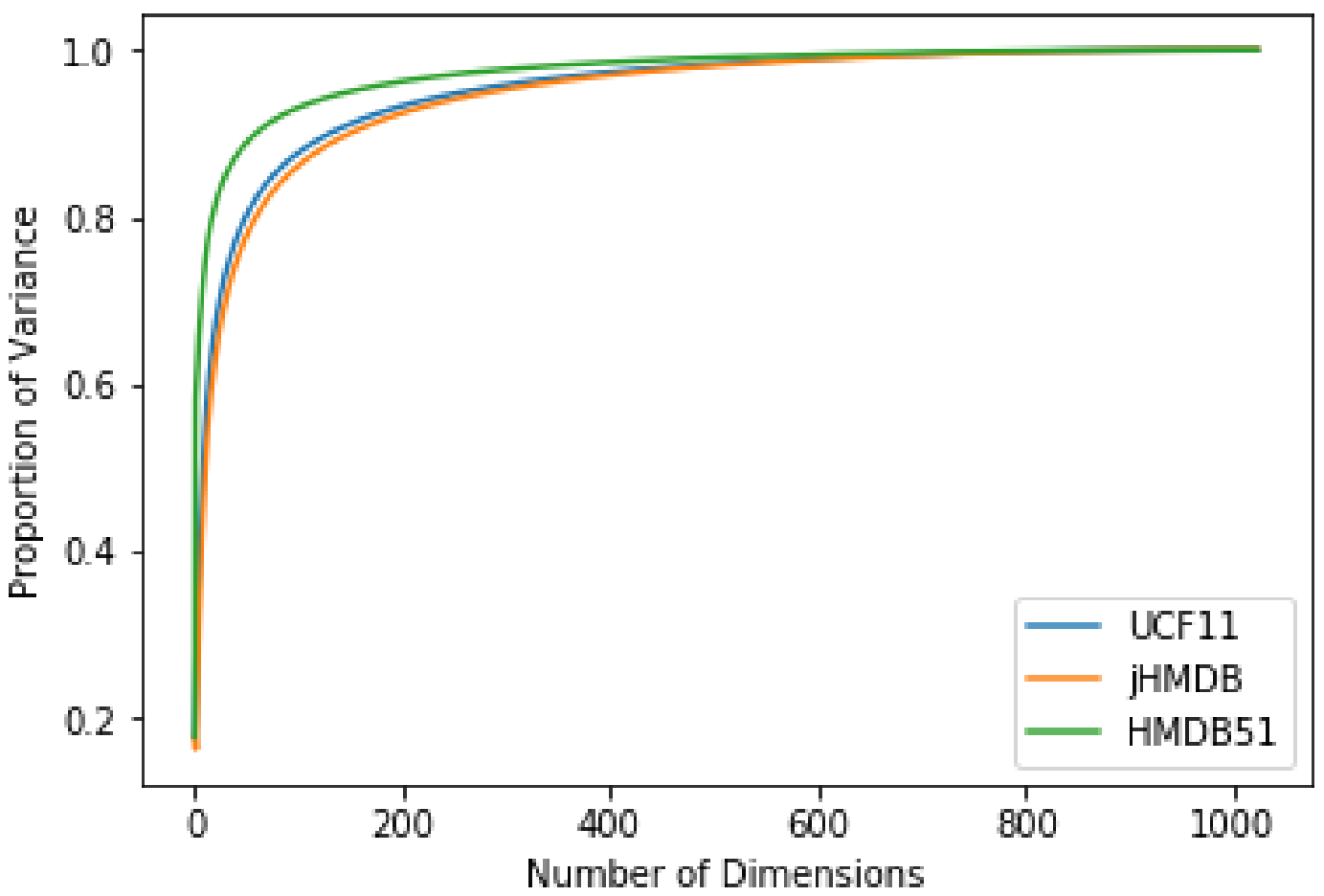

*Figure 6 PCA Energy Diagram for three datasets UCF11, jHMDB and HMDB51. Each value on the diagram represents the number of dimensions which corresponds to a specified POV value.*

*Table 1 Number of kept channels in each dataset*

| Network/Dataset | UCF11 | jHMDB | HMDB51 |
|---|---|---|---|
| **GoogLeNet** | 50 | 50 | 60 |
| **VGG-Net** | 60 | 60 | 65 |

### 4.2 Implementation Details

#### *4.2.1 Using Pre-trained CNN Networks to describe optical flow images*

For the description of optical flow images, two different pre-trained networks GoogLeNet and VGG-Net are used. These networks have already been trained on the ImageNet dataset. The representation of optical flow images is obtained via extracting the features of the last fully-connected layer (before the soft-max layer). GoogLeNet has 1024 neurons and VGG-Net has 4096 neurons in this layer.

#### *4.2.2 Principal Component Analysis setup*

As it was mentioned in the previous section, determining the Proportion of Variance (PoV) value is highly important as it specifies the number of channels that should be classified by the 1D-CNN. It should be determined by considering many factors e.g. the type of application to be used, the desired accuracy of the model and so on. Actually, there exists a trade-off for it. High values of Proportion of Variance (PoV) can add many redundant features which complicates the training the classifier; while, reducing it leads to losing useful information that can help the classifier for categorizing. Furthermore, since the next step is to train the Multi-Channel 1D-CNN, number of training data for training the 1D-CNN should also be considered carefully. Setting high values for the number of channels while not having enough training data leads to an unaccomplished trained CNN that will not work perfectly. On the other hand, reducing the number of channels to much low values, will restrict the capacity and the trainability of the proposed model.

Principal Component Analysis algorithm was performed on the optical flow descriptions of the three datasets. Fig. 6 shows the Proportion of Variance with respect to the number of channels for all training videos in datasets. As it is clear from the Fig. 6, because of the fact that the optical flow descriptions are highly correlated with each other, the number of dimensions can be reduced dramatically while keeping a reasonable amount of energy for all datasets. This point can make much more sense when considering HMDB51 since it has much more training videos than the other two datasets. As a matter of fact, the frame descriptions (for all datasets) are gathered at a part of space with huge number of dimensions; however, by performing PCA, we can describe them with much smaller dimensionalities. After a few experiments and also considering two important factors number of training videos and the PCA energy diagram (Fig. 6), the Proportion of Variance was set to 80% for UCF11 and jHMDB and for HMDB51 this value was set to 85%. Table. 1 indicates the number of channels in each dataset for two pre-trained CNN networks GoogLeNet and VGG-Net.

#### *4.2.3 Multi-Channel 1D-CNN architectures*

The proposed multi-channel one dimensional CNN architectures for the datasets can be viewed in Fig. 7. As it is indicated, the 1D-Module(*C,N,P*) is repeated inside the architectures. Due to the intrinsic differences in ranges of lengths and frame rates of the video clips, the characteristics of actions (e.g. the speed of activities) vary from one dataset to another. As a result, by doing a vast amount of experiments, we tried to find the optimal number of 1D-Modules with configuration settings of parameters *C, N, and P* in order to capture *sub-events* and consequently *super-events* of the ongoing actions within different datasets.

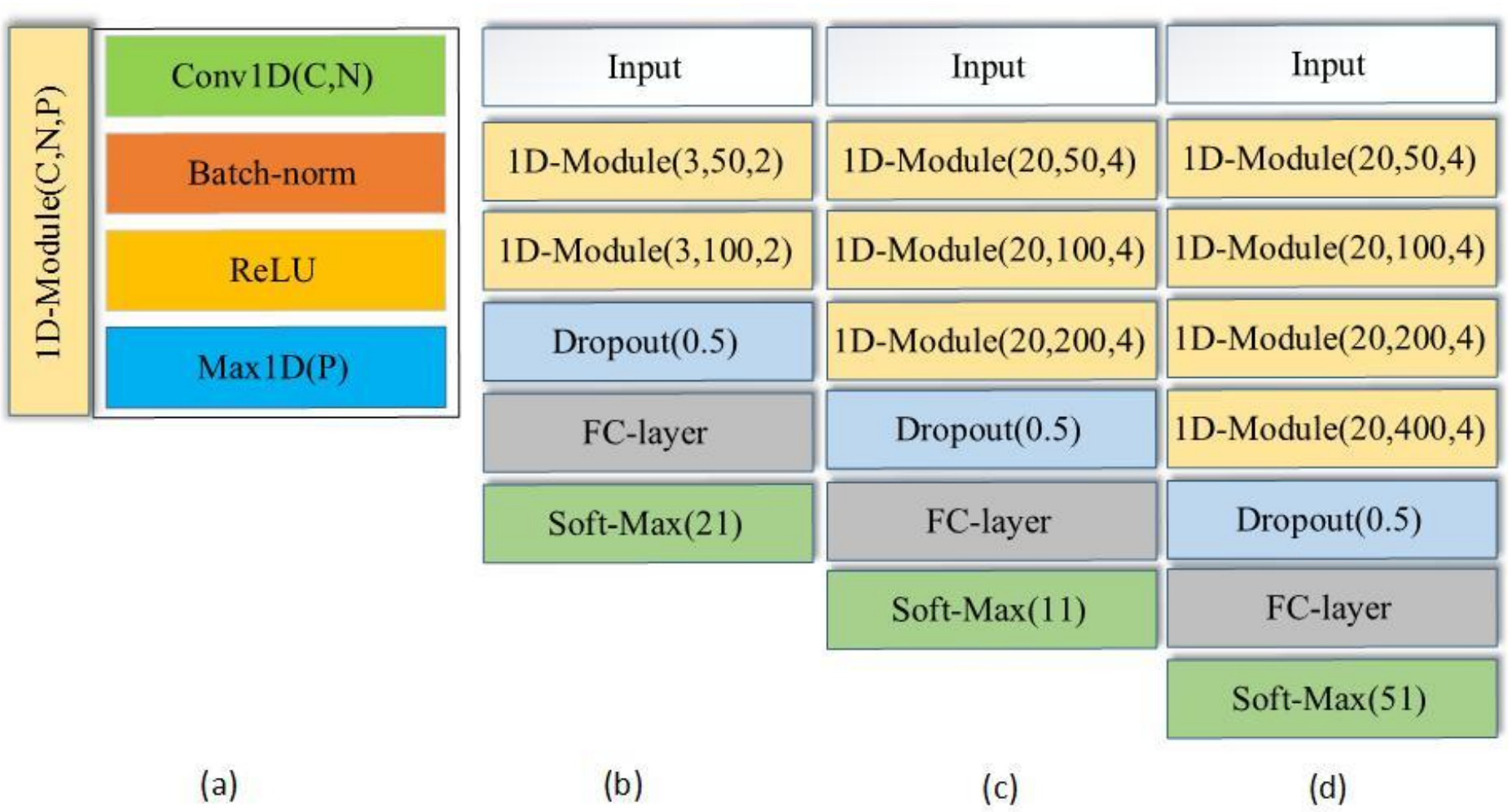

*Figure 7 The multi-channel one dimensional Convolutional Neural Network (1D-CNN) architectures used for three datasets jHMDB (b), UCF11 (c) and HMDB-51 (d). Inside 1D-Module is illustrated in (a). Conv1D (C,N) represents one dimensional convolution operator with C as the kernel size and N as the number of input channels. All 1D convolutions are operated with stride 1. Batch-norm layer standardizes the inputs and ReLU layer is the Rectified Linear Unit which acts as the activation function. Max1D(P) represents the one dimensional max pooling operator with a filter of size P. All 1D Max polling layers are operated with stride 2.*

## 4.3 Experimental Results

### *4.3.1 Results on UCF11*

To evaluate the proposed method on UCF11, the Leave-One-Out-Cross-Validation (LOOCV) scheme has been followed similar to the original work [26]. The results of classification accuracies and comparison with other state-of-the-art approaches are reported in Table 2. As it is clear, the proposed method has outperformed other methods.

*Table 2 Comparison of the proposed method with other state of the art methods on UCF11 dataset in terms of classification accuracy*

| Method | Accuracy(%) |
|---|---|
| Dense Trajectories [29] | 84.2% |
| Soft Attention [58] | 84.9% |
| DEEPEYE [59] | 86.1% |
| Cho *et al.* [60] | 88.0% |
| Chirag *et al.* [61] | 89.43% |
| Snippets [62] | 89.5% |
| Nazir *et al.* [63] | 93.9% |
| Two Stream LSTM (fu2) [47] | 94.6% |
| Ali *et al.* [56] | 95.7% |
| **Proposed method-Zero Padding (VGG-Net)** | **95.9%** |
| **Proposed method-Temporal Pyramid (VGG-Net)** | **96.1%** |
| Bag of Expression (BoE) [64] | 96.7% |
| **Proposed method-Zero Padding (GoogLeNet)** | **96.81%** |
| **Proposed method-Temporal Pyramid (GoogLeNet)** | **96.83%** |
| **Proposed method-Temporal Pyramid (GoogLeNet) + iTF** | **99.4%** |

*Table 3 Comparison of the proposed method with other state of the art methods on jHMDB dataset in terms of classification accuracy*

| Method | Accuracy(%) |
|---|---|
| PoTion [55] | 57.0% |
| P-CNN [65] | 61.1% |
| Action Tubes [66] | 62.5% |
| EleAtt-GRU [42] | 62.9% |
| **Proposed method-Zero Padding (VGG-Net)** | **68.8%** |
| PA3D [6] | 69.5% |
| **Proposed method-Temporal Pyramid (VGG-Net)** | **69.8%** |
| MR Two-Stream R-CNN [67] | 71.1% |
| DR$^2$N [68] | 71.8% |
| **Proposed method-Zero Padding (GoogLeNet)** | **72.3%** |
| Generalized Rank Pooling [69] + iTF | 73.7% |
| **Proposed method-Temporal Pyramid (GoogLeNet)** | **74.8%** |
| Chained MultiStream [70] | 76.1% |
| **Proposed method-Temporal Pyramid (GoogLeNet) + iTF** | **77.4%** |

#### *4.3.2 Results on jHMDB*

In the jHMDB dataset, evaluation is done regarding the three separate train and test splits suggested in the original dataset. The overall accuracy is the average performance of the splits. The classification accuracies of the proposed method and other approaches for this dataset is reported in Table 3. Despite the fact that jHMDB dataset is too small for training the multi-channel 1D-CNN, since it only has around 600 training videos, the proposed method is still demonstrating excellent results. By combining the learned features of our method with improved trajectory features (concatenating them); we could outperform the state-of-the-art approaches for this dataset.

#### *4.3.3 Results on HMDB51*

The evaluation in this dataset is the same as jHMDB since it has three splits and the mean classification accuracy is compared with other approaches. The recognition accuracies of the proposed method and other approaches for this dataset is reported in Table 4. Due to the fact that this dataset has much more training videos than the other two datasets, it is expected that the Multi-Channel 1D-CNN is trained superiorly. At the time designing the architecture of the 1D-CNN related to this dataset, we increased its complexity by adding more convolution and pooling layers. As a result, better classification accuracies were observed in compare with when the neural network had lesser complexities.

In all datasets, features of the last fully-connected layers are extracted as the representation of time series. These are the learnt features of the proposed methodology which are also concatenated with the improved Trajectory Features (iTF). Concatenating our 1D-CNN learnt features with iTF features leads to a better performance for all datasets. This remembers the point that these two types of features are complementary to each other. In fact, our 1D-CNN learns temporal features which are the result of convolution filters being performed on time series. Humans may not be able to interpret these learned features. While iTF method, by tracking the interest points through the video and describing these trajectories with human-defined descriptors, considers aspects of time series that the machine may not be able to detect. As a result, their combination constructs a strong representation for categorizing videos.

#### *4.3.4 Accuracy with respect to the number of channels*

We have also provided the recognition accuracies for the first splits of jHMDB and HMDB51 with respect to the number of channels after performing the dimension reduction algorithm. The results are obtained by using VGG-Net for describing optical flow frames. They can be seen in Fig. 8. The numbers are calculated by averaging the classification accuracies of our proposed algorithm after being run for 10 times. The standard deviation is also shown by vertical bars. As can be seen from Fig. 8, the mean classification accuracy is increased considerably for jHMDB, when the number of channels is raised up to 60. The same happens for HMDB51 when the number of channels is raised to 65. By increasing the number of channels for each dataset after the certain points mentioned above, both of them do not undergo significant changes in their accuracies. In fact, 60 channels for jHMDB and 65 channels for HMDB51 are the saturation points for them where by providing more information to the 1D-CNN, the neural network cannot achieve higher results. After that, by increasing the number of channels to values more than 70 for jHMDB and 80 for HMDB51, the accuracies decrease. This can be due to the fact that the 1D-CNN networks overfit and they are not trained perfectly.

*Table 4 Comparison of the proposed method with other state of the art methods on HMDB-51 dataset in terms of classification accuracy*

| Method | Accuracy(%) |
|---|---|
| PoTion [55] | 43.7% |
| C3D [5] | 51.6% |
| PA3D [6] | 55.3 |
| Two Stream [1] | 59.4% |
| iTF [28] | 61.7% |
| TDD [36] | 63.2% |
| **Proposed method-Zero Padding (VGG-Net)** | **65.3%** |
| **Proposed method-Temporal Pyramid (VGG-Net)** | **65.8%** |
| Generalized Rank Pooling [69] + iTF | 67% |
| **Proposed method-Zero Padding (GoogLeNet)** | **68.7%** |
| TSN [31] | 69.4% |
| **Proposed method-Temporal Pyramid (GoogLeNet)** | **69.5%** |
| IF-TTN [71] | 70% |
| ARTNet [72] | 70.9% |
| **Proposed method-Temporal Pyramid (GoogLeNet) + iTF** | **73.5%** |
| S3D [73] | 75.9% |

*4.3.5 Comparison of Number of Training Parameters*

In Table 5, the comparison of number of training parameters along with number of input frames between our 1D-CNN for HMDB51 and other methods is provided. As it is clear, our proposed method is able to reduce the number of training parameters by $\Theta(10^3)$. The reason for this dramatic decrease is converting the 3 dimensional video signals to multi-channel one dimensional time series; while other approaches such as C3D treat it as a 3 dimensional input and train a huge number of 3D filters to recognize the activities. The other superiority aspect is that the number of input frames in our method is not limited to a specific number. Due to the structure of either zero padding or temporal pyramid, the proposed method is able to handle videos with different lengths; whereas other approaches mentioned in the Table 5 do not have this capability. In fact, they are compelled to select between frames which may lose useful information. As a result, the proposed approach, by not being constrained to feed a specific number of frames to the network, not only demonstrates excellent results on multiple datasets, but also decreases the number of learning parameters significantly.

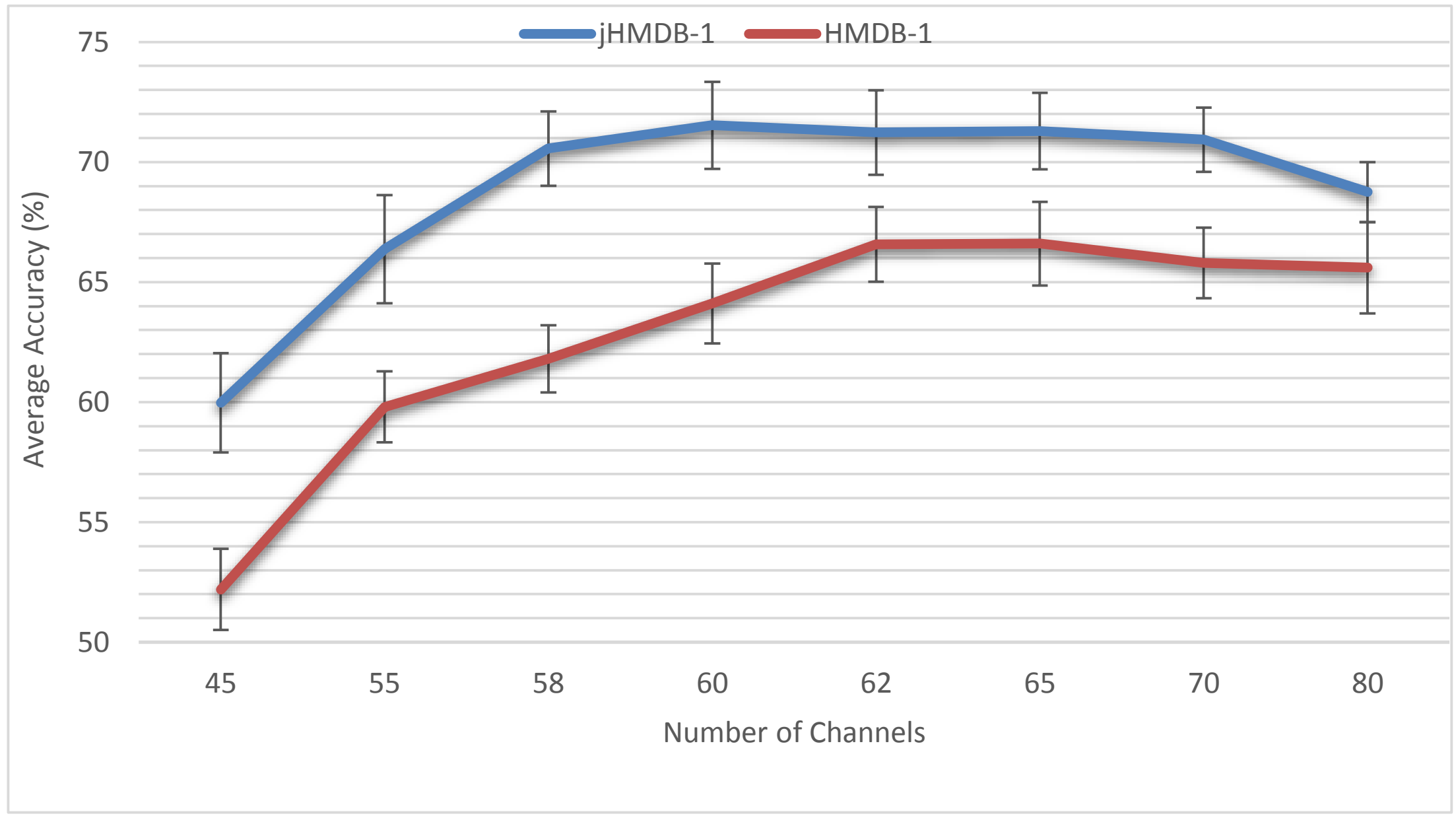

*Figure 8 The Average Accuracies with respect to the number of channels after performing PCA using VGG-Net for flow description.*

*Table 5 Comparison of number of training parameters and input frames between our 1D-CNN (HMDB-51) and other methods.*

| Method | Number of Parameters | Number of Input Frames |
|---|---|---|
| C3D [5] | 79M | 16 RGB |
| Two Stream (3D-Fused) [4] | 39M | 5 RGB, 50 Flow |
| Two-Stream I3D [54] | 25M | 64 RGB, 64 Flow |
| Two Stream [1] | 12M | 1 RGB, 10 Flow |
| ConvNet + LSTM [41] | 9M | 25 RGB |
| **Our 1D-CNN** | **50K** | **Not Limited** |

## 5. CONCLUSION

We proposed an approach in which a one dimensional Convolutional Neural Network (1D-CNN) in a multi-channel format is used to extract temporal information. By doing so, we showed that our 1D-CNN is able to extract both short-term and long-term motion dynamics due to its hierarchical structure. Furthermore, to control the number of learned parameters, the original feature space has been transformed to a less complicated space with a lower dimensionality. This is done by performing an effective PCA dimension reduction algorithm. In fact, by reducing the volume of input signals to a limited number of time series, the attention of our method is to capture motion changes across the whole video. By evaluating the proposed method on two public datasets UCF11 and jHMDB, it is demonstrated that our method could reach state-of-the-art results on both of them successfully. Furthermore, by comparing the results on jHMDB and HMDB51, it can be inferred that the more training data is provided for our deep learning neural network, the stronger accuracies for the classification model can be expected. As a result, the proposed method is trainable on small datasets which is a special property among most of the deep-learning based approaches. In future, we aim to provide an even lightened version of our model by doing extensive experiments with other architectures e.g. Fully Convolutional Networks (FCNs) which omits the last fully connected layers and replace them with convolutional layers. In this way, the neural network structure will become simpler since there will be no need to the Temporal Pyramid Pooling layer.